\documentclass[10pt,conference]{IEEEtran}

\usepackage{cite}

\usepackage{subcaption}
\usepackage{color}
\usepackage{booktabs} 
\usepackage{multirow}
\usepackage[normalem]{ulem}
\usepackage{amsmath,amssymb}

\ifCLASSINFOpdf
   \usepackage[pdftex]{graphicx}
   \graphicspath{{images/}}
   \DeclareGraphicsExtensions{.pdf,.jpeg,.png}
\else
   \usepackage[dvips]{graphicx}
   \graphicspath{{../images/}}
   \DeclareGraphicsExtensions{.eps}
\fi

\usepackage{algorithmic}

\usepackage{array}

\usepackage{url}

\newif\iffinal
\finaltrue

\iffinal
\else
\usepackage[switch]{lineno}
\fi

\begin{document}

%
\title{Recognizing Handwritten Mathematical Expressions of Vertical Addition and Subtraction}



\iffinal

\author{\IEEEauthorblockN{
Daniel Rosa\IEEEauthorrefmark{1},
Filipe R. Cordeiro\IEEEauthorrefmark{1},
Ruan Carvalho\IEEEauthorrefmark{1}, 
Everton Souza\IEEEauthorrefmark{1},
Sergio Chevtchenko\IEEEauthorrefmark{2},\\
Luiz Rodrigues\IEEEauthorrefmark{3},
Marcelo Marinho\IEEEauthorrefmark{1},
Thales Vieira\IEEEauthorrefmark{3} and
Valmir Macario\IEEEauthorrefmark{1}
}
\IEEEauthorblockA{\IEEEauthorrefmark{1} Visual Computing Lab, Department of Computing, 
Universidade Federal Rural de Pernambuco (UFRPE), Brazil}
\IEEEauthorblockA{\IEEEauthorrefmark{2}Centro de Inform\'atica - CIn, Universidade Federal de Pernambuco (UFPE), Brazil}
\IEEEauthorblockA{\IEEEauthorrefmark{3}Center for Excellence in Social Technologies (NEES), Federal University of Alagoas, Brazil}
Email: \{daniel.carneiro, filipe.rolim, ruan.carvalho, marcelo.marinho, valmir.macario\}@ufrpe.br, luiz.rodrigues@nees.ufal.br
}

\else
  \author{Sibgrapi paper ID: 6 \\ }
  \linenumbers
\fi

\maketitle
\let\thefootnote\relax\footnote{\\979-8-3503-3872-0/23/\$31.00
\textcopyright2023 IEEE}

\begin{abstract}

Handwritten Mathematical Expression Recognition (HMER) is a challenging task with many educational applications. Recent methods for HMER have been developed for complex mathematical expressions in standard horizontal format. However, solutions for elementary mathematical expression, such as vertical addition and subtraction, have not been explored in the literature. 
This work proposes a new handwritten elementary mathematical expression dataset composed of  addition and subtraction expressions in a vertical format. We also extended the MNIST dataset to generate artificial images with this structure. Furthermore, we  proposed a solution for offline HMER, able to recognize vertical addition and subtraction expressions. Our analysis evaluated the  object detection algorithms YOLO\_v7, YOLO\_v8, YOLO-NAS, NanoDet and FCOS for identifying the mathematical symbols. We also proposed a transcription method to map the bounding boxes from the object detection stage to a mathematical expression in the \LaTeX~markup sequence. Results show that our approach is efficient, achieving a high expression recognition rate. 
The code and dataset are available at \url{https://github.com/Danielgol/HME-VAS}
\end{abstract}


\IEEEpeerreviewmaketitle

\section{Introduction}
\label{sec:introduction}

Computer-Assisted Learning (CAL) tools can contribute positively to teaching and learning mathematics, helping with automatic examining scores and feedback \cite{meeter2021primary}.
A common task of these tools is to use an image of a handwritten problem, in which the system recognises the handwritten characters and solves the problem or gives feedback~\cite {neri2023methodology}. One of the main stages of CAL systems is the handwritten mathematical expression recognition (HMER), which is an image-to-text task to generate the corresponding mathematical symbols, usually in a  \LaTeX~markup sequence, from an input image. 

Although HMER can be compared to traditional handwriting text recognition tasks, it is considered more challenging~\cite{comer}\cite{zhelezniakov2021online}. One of the characteristics of HMER is the presence of spatial relations classification. In mathematical notation, spatial relationships are mostly used as implicit operators. For example, one number can be a superscript, subscript or regular number based on the expression's position, size and context. Besides, the incorrect classification of a single number or mathematical symbol changes the result of the operation and evaluation of the expression. Another challenge is the presence of complex structures and formulas and large amounts of symbols, often similar~\cite{beeton2001unicode}.

\begin{figure}[!ht]
\centering
\begin{subfigure}{0.5\columnwidth}
    \includegraphics[width=0.9\columnwidth]{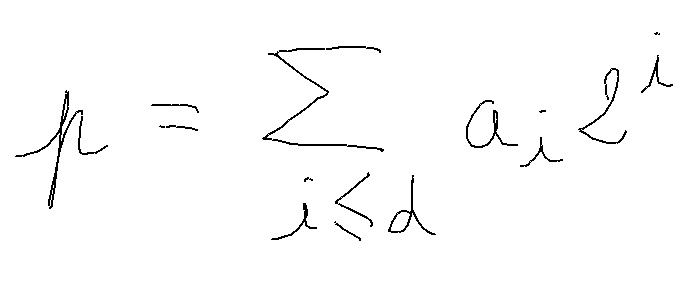}
    \caption{CROHME 2019} \label{fig:1a}
\end{subfigure}%
\hspace*{\fill}   
\begin{subfigure}{0.4\columnwidth}
\centering
\includegraphics[width=0.3\columnwidth]{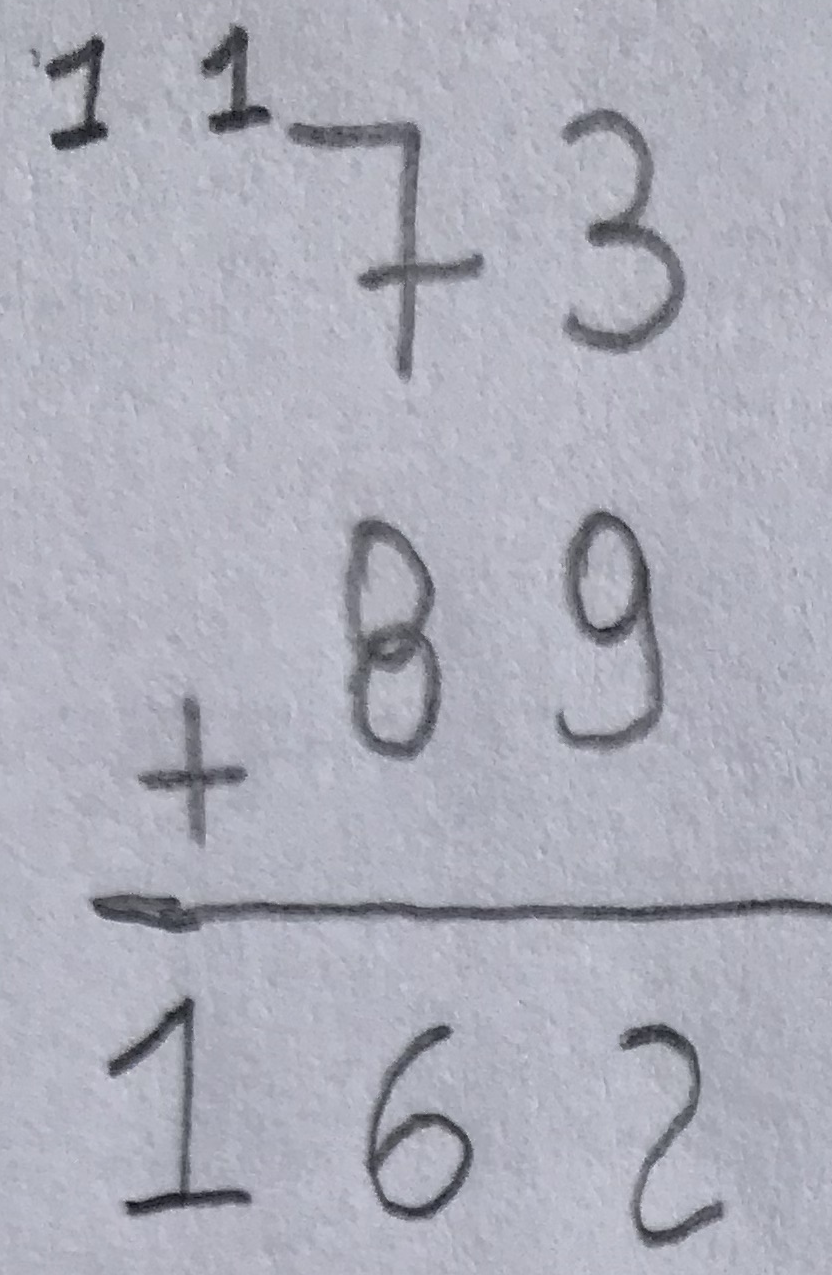}
 \caption{Our dataset} \label{fig:1b}
\end{subfigure}%
\caption{Image samples from (a) CROHME 2019~\cite{crohme19} and  (b) our built dataset. }
\label{fig:datasets_dif}
\end{figure}

Recent advancements in HMER have been proposed using deep learning models~\cite{bian2022handwritten}.
Most of the recent research in HMER uses popular datasets such as CROHME 2019~\cite{crohme19} and HME100k~\cite{hme100k}, which consist of thousands of images of complex equations in a horizontal format. However, none of these datasets contains equations in vertical format, which is used elementary mathematical expressions.  Vertical addition and subtraction use columns to align each number's place values, solving each one before combining for the solution. In this process, the numbers are lined up in columns according to their place values. The numbers are added or subtracted in each place value separately to calculate the solution. 
To the best of our knowledge, there is no work evaluating this type of equation format for HMER using deep learning models. Moreover, there is no public dataset of  mathematical expressions in vertical formats, such as column addition and subtraction. Fig.~\ref{fig:datasets_dif} shows images from CROHME, compared to our dataset. Although the images from our dataset are related to mathematical expressions, they are in a different domain because the equations are column-wise, there is a carry symbol in the expression, and they require a different training from the model.

In this work, we built an image dataset of basic addition and subtraction equations written in a vertical format. We also extended MNIST~\cite{mnist} dataset to generate artificial column addition and subtractions. We evaluated the object detection deep learning models  YOLO\_v7~\cite{yolov7}, YOLO\_v8~\cite{yolov8}, YOLO-NAS~\cite{yolonas}, NanoDet~\cite{nanodet} and FCOS~\cite{fcos}   for mathematical symbol detection and proposed a solution to HMER. The main contributions of this paper are as follows:

\begin{itemize}
\item  We propose a  dataset of handwritten column addition and subtraction format to simulate elementary school mathematical equations, which have not been explored in the literature. We also extended MNIST to generate images for the problem artificially;
\item  We evaluated state-of-the-art (SOTA) methods of object detection for  vertical addition and subtraction;
\item We propose a transcription stage to map the bounding boxes to a \LaTeX~expression, evaluating the ability of our solution to deal with the column-wise format.
\end{itemize}

\section{Related Work}
\label{sec:related_work}

Traditional HMER solutions divide the recognition task into three stages: (1) symbol recognition, (2) symbol classification and (3) structural analysis~\cite{chan2000mathematical}. Several strategies have been proposed to solve these problems sequentially, using Support Vector Machines~\cite{keshari2007hybrid}, Elastic Matching~\cite{vuong2010towards}
and tree transformation~\cite{zanibbi2002recognizing}. Hu and Zanibbi~\cite{hu2013segmenting} proposed a symbol segmentation method using AdaBoost algorithm and geometric
multi-scale shape context features. Le and Nakagawa~\cite{le2016system} used a SVM-based classifier for symbol segmentation, using 12 geometric
features and 9 additional features. Hu and Zanibbi~\cite{hu2016line} proposed a symbol segmentation algorithm based on graphs and Parzen window-modified Shape Context features. Fang and Zhang~\cite{fang2020multi} introduced a new approach to isolated symbol recognition called squeeze-extracted multi-feature convolution neural
network. In the structure analysis step, formal grammars have been proposed to recognize mathematical expression~\cite{lavirotte1998mathematical}\cite{chan2001error}.

In recent years, encoder-decoder architectures have shown increased performance in various image-to-text tasks, using deep learning models~\cite{bian2022handwritten}.
 Zhang et al.~\cite{zhang2017watch} were one of the first to use an  encoder-decoder neural architecture to solve HMER tasks with their model WAP, outperforming traditional grammar-based methods in the CROHME 2014 competition~\cite{crohme14}. Later, they proposed DenseWAP~\cite{zhang2018multi}, which uses a multi-scale DenseNet encoder to improve the ability to handle multi-scale symbols. Further, DenseWAP-TD~\cite{zhang2020tree} improves the ability to handle complex formulas by substituting a string decoder with a tree decoder. 
Although these end-to-end architectures perform well for horizontal handwritten mathematical expressions, they require large datasets to train the models and have millions of parameters.


Although SOTA methods in HMER have been extensively evaluated in large public datasets, such as CROHME 2019~\cite{crohme19} and HME100k~\cite{hme100k}, these datasets are composed of equations only in horizontal format. 
To the best of our knowledge, there is no public dataset from equations in vertical format, using column addition or subtraction, as shown in Fig~\ref{fig:datasets_dif}(b). Our work differs from previous  contributions by proposing an HMER solution for column-wise equation structure used in elementary school mathematics. We also simplify the structural analysis by using an algorithm based on the spatial relation of detected objects. Furthermore, we evaluated different SOTA backbones for our approach to this new task domain.

\section{Method}

\subsection{Problem Definition}
\label{sec:pd}

Consider $D=\{(x_1, y_1),...(x_n,y_n)\}$ as the training set of the handwritten mathematical expression recognition problem, where $\mathbf{x}_i \in \mathcal{X}$ is the $i^{th}$ image containing a mathematical expression in a vertical format and $\mathbf{y}_i \in \mathcal{Y}$ is the corresponding \LaTeX~expression.

A supervised classification learns a function $f: \mathcal{X} \to \mathcal{Y}$ that maps the input space $\mathcal{X}$ to the observed label space $\mathcal{Y}$. 
The function $f$ can be decomposed as  $f=g(h(\mathbf{x}_i))$, where 
$h:\mathcal{X} \to \mathcal{W}$ is an object detector that maps the input image 
$\mathbf{x}_i\in\mathcal{X}$ to a set $\mathbf{w}_i\in\mathbf{W}$ of classification and localization of mathematical 
symbols of $\mathbf{x}_i$, and $g:\mathcal{W}\to\mathcal{Y}$ is a transcriptor that maps the bounding boxes' positions and classes to a \LaTeX~expression. 
Therefore, $D$ can be extended to $\tilde{D}=\{(\mathbf{x}_n, \mathbf{w}_n, 
\mathbf{y}_n)\}_{i=1}^{n}$, where $\mathbf{w}_i=\{(\mathbf{l}_{i,j},\mathbf{b}_{i,j})\}_{j=1}^{|\mathbf{w}_i|}$ denotes the 
classification and localization of the $|w_i|$ mathematical symbols, 
with $\mathbf{l}_{i,j}$ denoting the label $l$ of the $j^{th}$ symbol and 
$\mathbf{b}_{i,j}\in \mathbb{R}^4$ representing the top-left and bottom-right coordinates of the bounding box of the $j^{th}$ symbol on $\mathbf{x}_i$.

We restrict the mathematical expressions in $\mathbf{x}_i$ as vertical sums and subtractions, using the operands ``$+$" and ``$-$". As in common addition and subtraction operations, it might contain carrying values, defined as the label $c$. Therefore, each bounding box has  an associated class $l\in\mathcal{L}$, where  $\mathcal{L}=\{0,1,2,3,4,5,6,7,8,9,+,-,=,c\}$. Although the carry symbol $c$ also is represented as the number 1 in the images, it depends on the size and positional features to be identified. 

The transcription labelling in $\mathcal{Y}$ is built through a  direct mapping of classes  in $\mathcal{L}$ to the \LaTeX~expression, except for the presence of carrying symbols,  which we use ``$\backslash$overset\{\}\{\}"  expression. For example, the number 2 with a carry is denoted as ``$\backslash$overset\{1\}\{2\}" in the \LaTeX~markup sequence.


\subsection{Proposed Solution}

\begin{figure*}[!ht]
\centering
\includegraphics[ width=1.8\columnwidth]{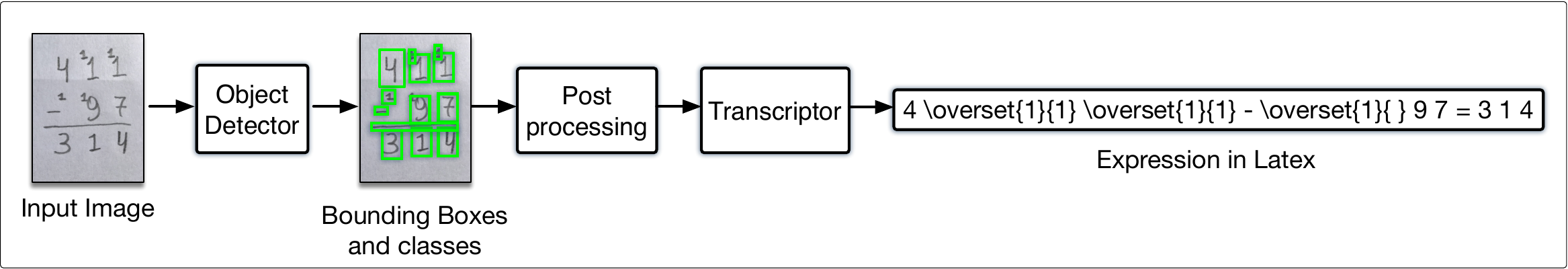}

\caption{Flowchart of the proposed solution for column addition and subtraction recognition.}
\label{fig:flowchart}
\end{figure*}

The proposed solution divides the expression recognition problem into two main stages: (1) Object Detection and (2) Transcription. In the object detection stage, we aim to identify the localization of the mathematical symbols and obtain the corresponding bounding boxes. Next, we perform a post-processing step to improve the detection results. Once we have the position of each mathematical symbol in the image, we use a transcriptor that receives as input the bounding box and classes and produces a resulting expression in \LaTeX.
Fig.~\ref{fig:flowchart} shows the proposed HMER solution.

\subsubsection{Object Detection}

In the object detection stage, we apply different state-of-the-art methods based on convolutional neural networks. For this stage, we evaluated the models YOLO\_v7~\cite{yolov7}, YOLO\_v8~\cite{yolov8}, YOLO\_NAS~\cite{yolonas}, FCOS~\cite{fcos}  and NanoDet~\cite{nanodet}. All these approaches receive as input an RGB image and produce as output the bounding boxes and corresponding classes. 

We perform a post-processing stage to improve the obtained detection. In this stage, we first filter the bounding boxes with a confidence level $\zeta$ lower than a threshold $\theta$, removing them. In the second step, when two or more bounding boxes have an Intersection Over Union (IoU) value $IoU>\alpha$, we keep the bounding box of the most confident class and discard the remaining ones. The values of $\theta$ and $\alpha$ are selected using an optimizer on the validation set. This strategy showed to improve the detection and removed duplicated bounding boxes. 



\subsubsection{Transcription}

\begin{figure}[!ht]
\centering
\includegraphics[ width=0.9\columnwidth]{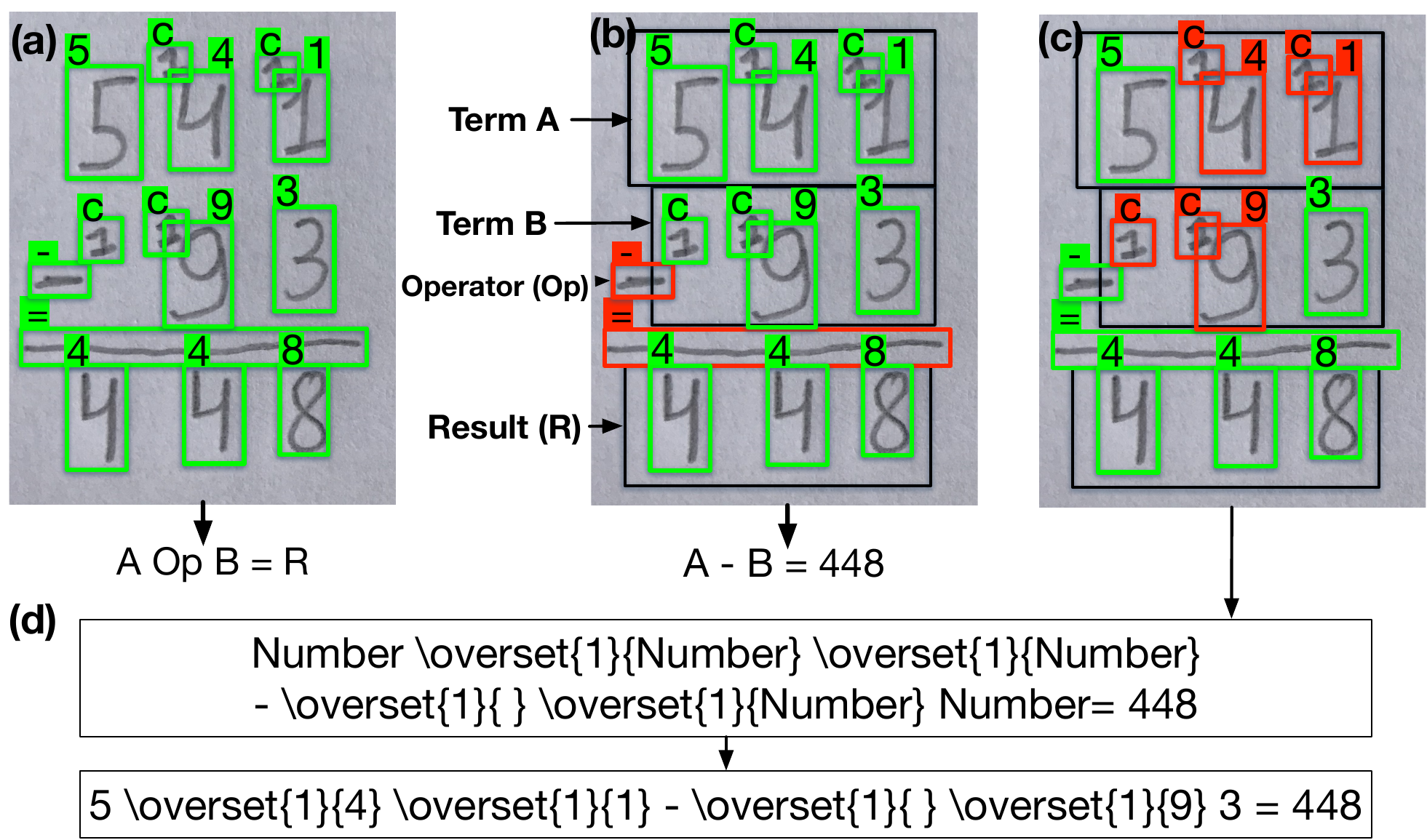}

\caption{Transcription stage. (a) bounding boxes and predictions of the object detection; (b) Identification of operator \textit{Op}, equal symbol and result numbers. The results are obtained from the number below equal sign's bounding box;  (c) identification of carry symbols and number associated, and (d) construction of the expression based on the identified bounding boxes.}
\label{fig:transcription}
\end{figure}

The transcription stage receives the bounding boxes and corresponding classes as input and produces a \LaTeX~expression related to the mathematical equation in the image. Some traditional approaches for HMER solve this problem by either building a grammar~\cite{alvaro2016integrated} or using a graph strategy~\cite{lods2019fuzzy}. In this work, we simplify this stage by proposing a transcription based on the positional structure of the bounding boxes. Fig.~\ref{fig:transcription} shows each step of the proposed transcription stage, described below.

At the transcriptor stage, the detections are divided into four categories: number, operation (op), equal sign, and carry, where the number refers to the digits 0 to 9, operation $\in \{+,-\}$, equal sign refers to the equal line in the expression, and carry refers to the small ``1" digits related to the carry symbol.
The expression is expected to be in the format ``A op B = R", where A and B are the operands and R is the resulting number. The operation and equal symbols are both expected to have only one bounding box associated with each image while
the numbers and carry categories are expected to have varying amounts of detections. 

Based on the bounding box position of the equals sign, it is possible to define the regions of the resulting term and the operand terms of the equation. The bounding boxes below the equal sign are related to R, while the others are associated with the operands A or B. 
From the resulting term R, we define its transcription based on the predicted class since it is composed only of numbers and has no carry symbol. The numbers of R are built based on the $x$ coordinates of the bounding boxes in R, from left to right.
The next step is identifying the bounding boxes related to the expression's operands, including the associated carry symbols. To do so, we group the remaining bounding boxes (i.e. not include operator, equal sign and resulting term) in two groups, based on the $y$ value of the top-left $(x,y)$ coordinate of each bounding box, defining the elements of operands A and B. For each term, we identify the numbers and carry symbols. 

In order to associate each  carry symbol with its respective number, another sequence is followed: First, the digits from each term are taken and checked if there is any  carry symbol nearby. This check occurs from the rightmost to the leftmost digit from the actual term, starting from \textit{A} term to \textit{B}. 
 We associate a carry symbol with a related digit based on the intersection of the carry and digit candidate.  


Once the carry is associated with a specific digit, it can no longer be related to any other digit. If any carry has not been linked to any digit, it will be considered isolated and unrelated to any number.
All information needed to build the equation \LaTeX~expression is extracted at the end of the mentioned processes. To generate it, it is necessary to translate each digit of the following data from left to right.  All the numbers associated with the carry values are represented in  \LaTeX~as ``$\backslash$overset", as described in Sec.~\ref{sec:pd}. 

\section{Methodology}

\subsection{Dataset}
\label{sec:dataset}

\begin{figure}[!ht]
\centering
\includegraphics[ width=\columnwidth]{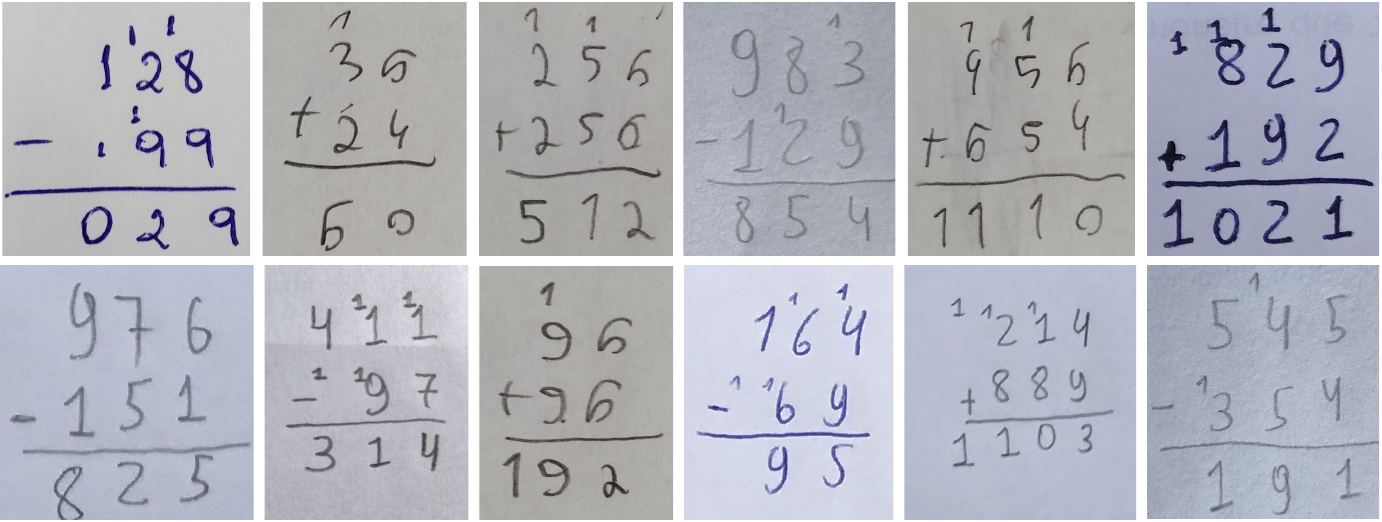}

\caption{Examples of images from our proposed column addition and subtraction dataset. }
\label{fig:our_dataset}
\end{figure}

We built a custom dataset of elemental addition and subtraction arithmetic equations written in a vertical format. Fig.~\ref{fig:our_dataset} shows image samples from our built dataset. The dataset comprises 300 images, with mathematical expressions written by four human annotators and acquired using a camera. The images from the dataset vary in dimension, but all are resized to $320\times320$ pixels to train the model. Besides the different writing styles, image writing also varies using a pencil or pen.  The ground truth of each image is saved in a \LaTeX~format.  

We considered 14 classes for the proposed dataset, composed of 10 classes related to the digits from 0 to 9, 2 classes corresponding to the symbols of sum and subtraction, 1 class related to the equal symbol, and 1 class corresponding to the carry symbol. Although the carry has the same shape representation as the digit $1$, the carry is also identified by its size related to the other numbers and its position. 

To define the training and test sets, we use a 3-fold cross-validation approach, where at each evaluation, we leave one annotator out as the test set and use the remaining ones as the training set. We refer to the annotators as H1, H2, H3 and H4.  We define the following splits in the format ``training set/ test set": (1) training set: H1, H2, and H3/  test set: H4  ; (2) training set: H1, H3 and H4/ test set: H2, (3) training set: H1, H2 and H4/ test set: H3. Table~\ref{tab:dataset_info} shows the dataset distribution. For each training split, we separate 50 images from the training set as the validation set.

\begin{table}[ht]
\caption{Our dataset distribution.}
\label{tab:dataset_info}
\centering
\begin{tabular}{ccccc}
\toprule
 & Number of Images & Split 1 & Split 2 & Split 3 \\
 \midrule
Annotator 1 (H1) &  151 & Train & Train & Train\\
Annotator 2 (H2) & 75 & Train & Train & Test\\
Annotator 3 (H3) & 50 & Train & Test & Train\\
Annotator 4 (H4) & 25 & Test & Train & Train\\
\midrule
Total & 300 \\
\bottomrule
\end{tabular}
\end{table}


\begin{figure}[!ht]
\centering
\includegraphics[ width=0.9\columnwidth]{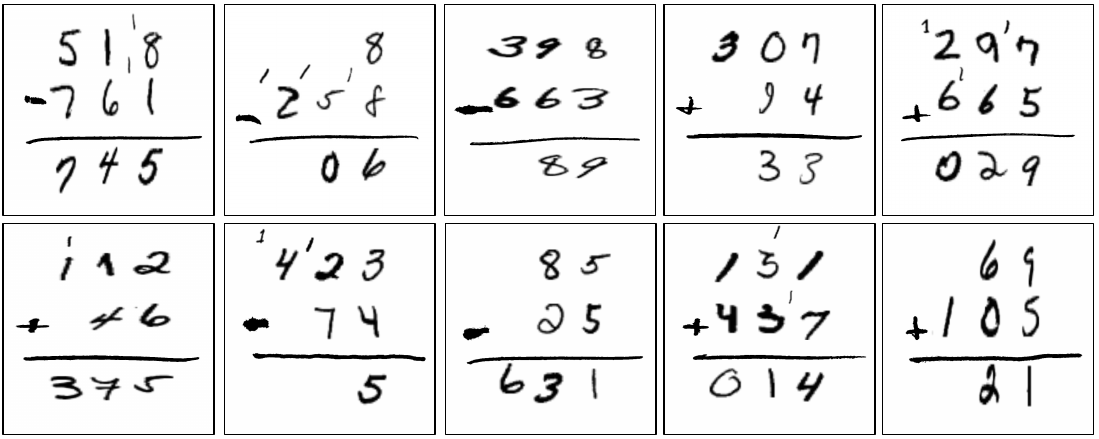}

\caption{Extended MNIST dataset based on mathematical expressions using MNIST~\cite{mnist} digits.}
\label{fig:mnist}
\end{figure}

We also proposed a dataset augmentation based on MNIST~\cite{mnist} dataset. MNIST consists of handwritten digit images of size $28 \times 28$ pixels with a training set of 60,000 samples and a test set of 10,000 samples. As the MNIST dataset contains only one digit by image, with no mathematical expression, we built new images using the position of the mathematical expressions of our dataset but replacing the number for the MNIST digits. We also used symbols ``+", ``-", and ``=" from our dataset but used the images from MNIST to fill the position of the digits randomly. 
For this augmented dataset, we did not care about the correctness of the equation because our goal is to recognize the mathematical expression independent of the correctness. Fig.~\ref{fig:mnist} shows examples of the extended MNIST dataset. This artificially extended dataset is composed of 2,600 images using MNIST digits. 





In our work, we combined our  300 handwritten images dataset  with the 2,600 augmented MNIST for the training set.  The test and validation sets are  composed only of the built handwritten images with human annotators.

\subsection{Metrics}
\label{sec:metrics}

We evaluated the proposed solution and different object detection methods by using metrics related to the quality of the detection and transcription.

Average Precision (AP) is a commonly used metric for object detection, derived from precision and recall. Average precision computes the average precision value for recall value over 0 to 1 and is usually evaluated separately for each object category. Precision is derived from Intersection over Union (IoU), which is the ratio of the area of overlap and the area of union between the ground truth and the predicted bounding box. A threshold is set to determine if the detection is correct.
Precision measures the percentage of correct predictions, while recall measures the correct predictions to the ground truth. 
To compare performance overall object categories, the mean AP (mAP) averaged over all object categories is adopted as the final measure of performance.

Expression recognition rate (ER), defined as the percentage of correctly recognized expressions, is used to evaluate the performance of different methods on mathematical expression recognition~\cite{comer}\cite{li2022counting}. Moreover, $\leq1$
and $\leq2$ are also used, indicating that the expression recognition rate is tolerable at most one or two symbol-level errors. ER can be defined as:

\begin{equation}
    \label{eq:ER}
    ER = \frac{\text{Number of correctly recognized expressions}}{\text{Total number of expressions}}
\end{equation}

\subsection{Implementation}

 For all models, we initialized the models using the pre-trained weights, pre-trained at COCO dataset~\cite{coco}. For all models, we used their implementation in the provided repositories with the default parameters. We resized all the images from our dataset to 320$\times$320 pixels and trained all the models for 300 epochs. All the images are in RGB colour space. We used Optuna~\cite{optuna} to optimize the parameters $\alpha$ and $
 \theta$ in the post-processing stage, using the validation set. The models were trained and evaluated using the splits defined in Tab.~\ref{tab:dataset_info}, but we also added the 2600 artificial images from the extended MNIST in each training split. The validation and test sets are composed only of images from human annotators.

\section{Results and Discussion}
\label{sec:results}

We evaluated our solution using different object detectors for the proposed dataset: YOLO\_v7~\cite{yolov7}, YOLO\_v8~\cite{yolov8}, YOLO-NAS~\cite{yolonas}, NanoDet~\cite{nanodet} and FCOS~\cite{fcos}. We selected these models because they are SOTA for object detection~\cite{arani2022comprehensive}.  We divided our analysis into three subsections: Object Detection, Ablation Study and Expression Recognition. 

\subsection{Object Detection}

We evaluated the best object detector for the task of detecting the handwritten mathematical symbols, as defined in Sec.~\ref{sec:dataset}. Tab.~\ref{tab:res_od} shows the mAP results, considering the different annotators as test set (H2, H3, H4) and the average results.

\begin{table}[ht]
\caption{Results of mAP for the object detection stage, reported @IoU=0.5. H2, H3 and H3 denote different annotators for the test set. The top methods are in bold.}
\label{tab:res_od}
\centering
\begin{tabular}{c|c|c|c|c}
\toprule
Method & H2 & H3 & H4 & Avg. \\
\midrule
YOLO\_v7~\cite{yolov7} & \text{87.29} & \text{98.07} & \text{99.55} & 94.97 \\
YOLO\_v8~\cite{yolov8} & \textbf{87.64} & \textbf{99.68}  & \text{98.20} & \textbf{95.17} \\
YOLO-NAS~\cite{yolonas} & \text{66.81} & \text{74.63}  & \textbf{99.67} & 80.37 \\
NanoDet~\cite{nanodet}  & \text{82.89} & \text{98.62}  & \text{99.32} & 93.61 \\
FCOS~\cite{fcos} & \text{81.34} & \text{98.51}  & \text{95.46} & 91.77 \\
 \bottomrule
\end{tabular}

\end{table}

\begin{figure}[!ht]
\centering
\includegraphics[ width=0.9\columnwidth]{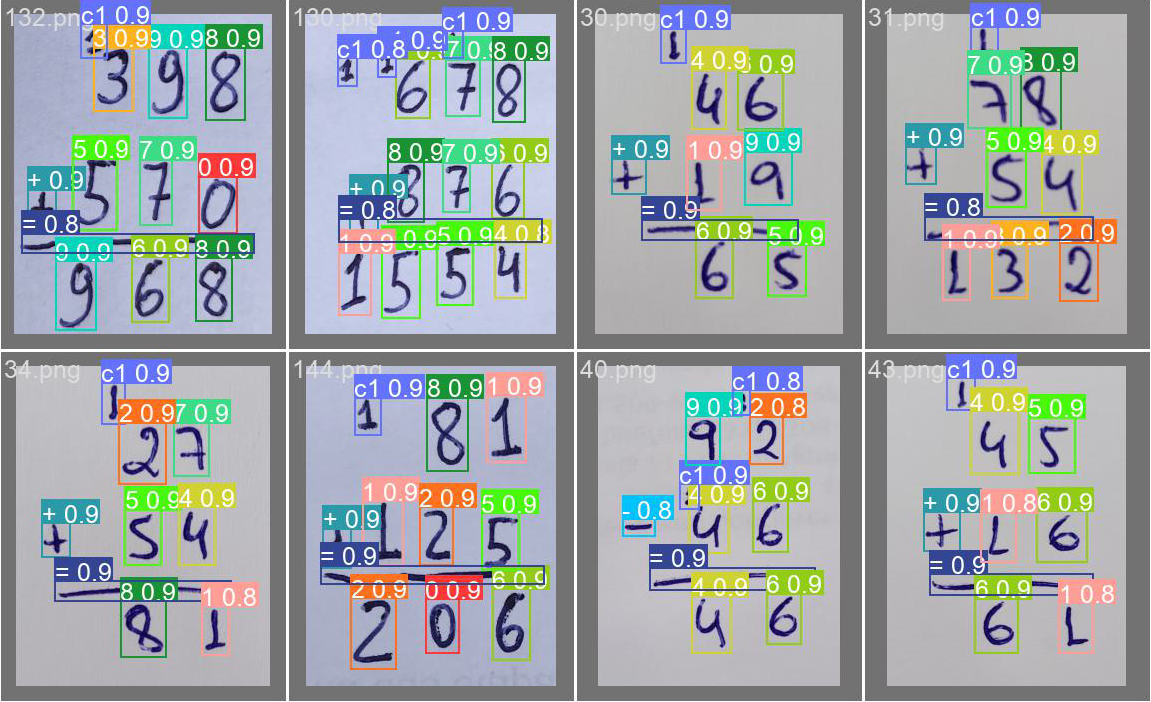}

\caption{Results of object detection of the mathematical symbols for  vertical addition and subtraction. }
\label{fig:detection_res}
\end{figure}

Results show that YOLO\_v8 presented the best results for object recognition compared with the other state-of-the-art methods. The annotator H2 has the most challenging symbols for the models to detect  because many symbols are written similarly. Although YOLO\_v7 also had high accuracy, on average, YOLO\_v8 had the best results. Fig.~\ref{fig:detection_res} shows examples of the detection of this method, which indicates the high quality of the detection, despite the small-size training set. 

\subsection{Ablation Study}

We also evaluated the detection quality based on the built training set and the extended MNIST. For this purpose, we evaluated the different training set scenarios: (1) using only our proposed small-training dataset; (2) using only the artificial equations based on augmented MNIST, and (3) using a combination of augmented MNIST and the proposed dataset.
For all scenarios, the validation and test sets are composed only of our handwritten images. 
Table~\ref{tab:ablation} shows results considering these different setups, using YOLO\_v8 object detector.

\begin{table}[ht]
\caption{Ablation study using our handwritten dataset and the augmented  MNIST, using YOLO\_v8 object detector. The top methods are in bold.}
\label{tab:ablation}
\centering
\begin{tabular}{c|c|c|c|c}
\toprule
Training Data & H2 & H3 & H4 & Avg. \\
\midrule
Our    & 80.39 & 96.05 & \textbf{98.32} & 91.58\\
Augmented MNIST   &  50.75 & 84.29 & 76.59 & 70.54 \\
Our + Augmented MNIST   &  \textbf{87.64} & \textbf{99.68} & 98.20 & \textbf{95.17} \\
 \bottomrule
\end{tabular}

\end{table}

Results show that the combination ``Our + Augmented MNIST " produces the best results than only using MNIST-based expressions or our handwritten dataset. This shows that using artificial images generated from MNIST digits, combined with real handwritten images, improves the results, and both datasets are important to solve the problem.  

\subsection{Expression Recognition}

We analyzed the expression recognition of the proposed method using different methods for object detection. We used the metric ER for this task, defined in Sec.~\ref{sec:metrics}. Table~\ref{tab:res_er} shows the results for ER, where $\leq1$ and $\leq2$ denote an error tolerance of 1 and 2 symbols, respectively. 

\begin{table*}[ht]
\caption{Results of Expression Recognition Rate (ER).  ER, “$\leq1$ error” and “$\leq2$ error” columns mean expression recognition rate when zero to two structural or symbol errors can be tolerated. All results are reported as a percentage (\%). H2, H3 and H4 denote different annotators
for the test set. The top methods are in bold.}
\label{tab:res_er}
\centering
\scalebox{0.90}{
\begin{tabular}{c|ccc|ccc|ccc|ccc}
\toprule
\multirow{2}{*}{Method} & \multicolumn{3}{c|}{H2} & \multicolumn{3}{c|}{H3} & \multicolumn{3}{c|}{H4} & \multicolumn{3}{c}{Avg} \\
 & ER & $\leq1$ error & $\leq2$ error & ER & $\leq1$ error & $\leq2$ errors 
 & ER & $\leq1$ error & $\leq2$ error & ER & $\leq1$ error & $\leq2$ error\\
\midrule
YOLO\_v7~\cite{yolov7} & \text{44.28} & \text{71.42} & \text{88.57} & \text{72.00} & \text{96.00} & \textbf{100} & \textbf{100} & \textbf{100} & \textbf{100} & 72.09 & 89.14 & 96.19 \\
YOLO\_v8~\cite{yolov8} & \textbf{48.57} & \textbf{75.71} & \textbf{91.42} & \textbf{96.00} & \textbf{100} & \textbf{100} & \text{84.00} & \text{96.00} & \textbf{100} & \textbf{76.19} & \textbf{90.57} & \textbf{97.14} \\
YOLO-NAS~\cite{yolonas} & \text{12.85} & \text{28.57} & \text{47.14} & \text{8.00} & \text{26.00} & \text{54.00} & \text{92.00} & \textbf{100} & \textbf{100} & 37.61 & 51.52 & 67.04  \\
NanoDet~\cite{nanodet}  & \text{22.85} & \text{47.14} & \text{71.42} & \text{84.00} & \text{98.00} & \textbf{100} & \text{92.00} & \textbf{100} & \textbf{100} & 66.28 & 81.71 & 90.47  \\
FCOS~\cite{fcos}  & \text{27.14} & \text{52.85} & \text{81.42} & \text{90.00} & \textbf{100} & \textbf{100} & \text{68.00} & \text{88.00} & \text{96.00} & 61.71 & 80.28 & 92.47  \\
 
 \bottomrule
\end{tabular}
}

\end{table*}

Results show that method YOLO\_v8,  combined with our proposed transcription strategy, obtained the best results with an ER of 76.19\% on average. As the H2 annotator had the hardest digits to detect, as shown in Tab ~\ref{tab:res_od}, it impacted the ER metric, as each misclassified symbol counts as an error for the expression. This shows that the area and the built dataset are challenging, and the proposed solution is a promising direction to deal with this type of equation.


\section{Conclusion}
\label{sec:conclusion}

In this work, we proposed a solution for automatically detecting and recognising handwritten mathematical expressions in vertical format, which is used for basic mathematical teaching. As there is no public dataset for this task, we built a dataset and developed a method for identifying the mathematical expression. We also extended MNIST to generate vertical addition and subtractions, improving the results.

    We also evaluated  SOTA object detection methods for identifying mathematical symbols, and  YOLO\_v8 obtained the best results, with 95.17\% mAP on average. We also proposed a transcription stage to transform the bounding boxes in a \LaTeX~expression and could obtain high-quality results in expression recognition metric.

This work provided a simple and effective solution to identify HME in a vertical format, which was not explored in the literature. In future works, we aim to expand the number of symbols and complexity of the expressions.



\bibliographystyle{IEEEtran}
\bibliography{example}
%
%


\end{document}